\pdfoutput=1

\documentclass[11pt]{article}

\usepackage[]{emnlp2021}

\usepackage{times}
\usepackage{latexsym}
\usepackage{booktabs}
\usepackage[T1]{fontenc}
\usepackage[linesnumbered,ruled]{algorithm2e}

\usepackage[utf8]{inputenc}
\usepackage{microtype}
\usepackage{amsmath}
\usepackage{amssymb}
\usepackage{graphics}
\usepackage{multirow}
\usepackage{tabularx}
\usepackage{colortbl}
\usepackage{subcaption}
 \usepackage{graphicx}
 \usepackage{hyperref}
\newcolumntype{s}{>{\columncolor[gray]{.85}[.5\tabcolsep]}c}

\title{Improving Multilingual Translation by Representation and Gradient Regularization}

\author{
    Yilin Yang$^{1, }$\thanks{\ \ Work done while interning at Microsoft.} \ ,
    Akiko Eriguchi$^2$,
    Alexandre Muzio$^2$, \\
    \textbf{Prasad Tadepalli}$^1$,
    \textbf{Stefan Lee}$^1$,
    \textbf{Hany Hassan}$^2$ \\
    $^1$Oregon State University \\
    $^2$Microsoft \\
    $^1$\texttt{\{yangyil,tadepall,leestef\}@oregonstate.edu} \\
    $^2$\texttt{\{akikoe,alferre,hanyh\}@microsoft.com}
}

\begin{document}
\maketitle
\begin{abstract}
Multilingual Neural Machine Translation (NMT) enables one model to serve all translation directions, including ones that are unseen during training, i.e. zero-shot translation. Despite being theoretically attractive, current models often produce low quality translations -- commonly failing to even produce outputs in the right target language. In this work, we observe that off-target translation is dominant even in strong multilingual systems, trained on massive multilingual corpora. To address this issue, we propose a joint approach to regularize NMT models at both representation-level and gradient-level. At the representation level, we leverage an auxiliary target language prediction task to regularize decoder outputs to retain information about the target language. At the gradient level, we leverage a small amount of direct data (in thousands of sentence pairs) to regularize model gradients. Our results demonstrate that our approach is highly effective in both reducing off-target translation occurrences and improving zero-shot translation performance by +5.59 and +10.38 BLEU on WMT and OPUS datasets respectively. Moreover, experiments show that our method also works well when the small amount of direct data is not available.\footnote{Codes and rebuilt OPUS data at: \url{https://github.com/yilinyang7/fairseq_multi_fix}}
\end{abstract}

\section{Introduction}
With Neural Machine Translation becoming the state-of-the-art approach in bilingual machine translations~\cite{bahdanau2014neural,vaswani2017attention}, Multilingual NMT systems have increasingly  gained  attention due to their deployment efficiency.
One conceptually attractive advantage of Multilingual NMT~\cite{johnson2017google} is its capability to translate between multiple source and target languages with only one model, where many directions\footnote{Most NMT dataset are English-centric, meaning that all training pairs include English either as source or target.} are trained in a zero-shot manner.

Despite its theoretical benefits, Multilingual NMT often suffers from \textit{target language interference}~\cite{johnson2017google,wang2020negative}. Specifically, \citet{johnson2017google} found that Multilingual NMT often improves performance compared to bilingual models in many-to-one setting (translating other languages into English), yet often hurts performance in one-to-many setting (translating English into other languages).
Several other works~\cite{wang2018three,arivazhagan2019massively,tang2020multilingual} also confirm one-to-many translation to be more challenging than many-to-one.
Another widely observed phenomenon is that the current multilingual system on zero-shot translations faces serious \emph{off-target translation} issue~\cite{gu2019improved,zhang2020improving} where the generated target text is not in the intended language.
For example, Table~\ref{tab:offtarget} shows the percentage of off-target translations appearing between high-resource languages.
These issues exemplify the internal failure of multilingual systems to model different target languages.
This paper focuses on reducing off-target translation, since it has the potential to improve the quality of zero-shot translation as well as general translation accuracy. 

\begin{table}[t]
    \centering
    \scalebox{0.9}{
        \begin{tabular}{c|cccc}
            \toprule
            WMT & Fr-De & De-Fr & Cs-De & De-Cs \\
            Baseline & 51.60\% & 39.80\% & 13.10\% & 20.50\% \\
            \midrule
            OPUS & Fr-De & De-Fr & Fr-Ru & Ru-Fr \\
            Baseline & 95.15\% & 93.70\% & 68.85\% & 91.20\% \\
            \bottomrule
        \end{tabular}
    }
    \caption{Off-target translation percentages on WMT and OPUS Testsets.}
    \label{tab:offtarget}
\end{table}

Previous work on reducing the off-target issue often resorts to back-translation (BT) techniques~\cite{sennrich2015improving}.
\citet{gu2019improved} employs a pretrained NMT model to generate BT parallel data for all $O(N^2)$ English-free\footnote{We denote translation directions that do not involve English
as \emph{English-free} directions.} directions and trains the multilingual systems on both real and synthetic data.
\citet{zhang2020improving} instead fine-tune the pretrained multilingual system on BT data that are
randomly
generated online for all zero-shot directions.
However, leveraging BT data for zero-shot directions has some weaknesses:
\begin{itemize}
    \item The need for BT data grows quadratically with the number of languages involved, requiring significant time and computing resources to generate the synthetic data.
    \item Training the multilingual systems on noisy BT data would usually hurt the English-centric performance~\cite{zhang2020improving}.
\end{itemize}
In this work, we propose a joint representation-level and gradient-level regularization to directly address multilingual system's limitation
of modeling different target languages.
At representation-level, we regulate the NMT decoder states by adding an auxiliary Target Language Prediction (TLP) loss, such that decoder outputs are retained with target language information.
At gradient-level, we leverage a small amount of direct data (in thousands of sentence pairs) to project the model gradients for each target language (TGP for Target-Gradient-Projection).
We evaluate our methods on two large scale datasets, one concatenated from previous WMT competitions with 10 languages, and the OPUS-100 from ~\citet{zhang2020improving} with 95  languages.
Our results demonstrate the effectiveness of our approaches in all language pairs, with an average +5.59 and +10.38 BLEU gain across zero-shot pairs, and 24.5\% $\to$ 0.9\% and 65.8\% $\to$ 4.7\% reduction to off-target rates on WMT-10 and OPUS-100 respectively.
Moreover, we show the off-target translation not only appears in the zero-shot directions, but also exists in the English-centric pairs.

\section{Approach}
In this section, we will illustrate the baseline multilingual models and our proposed joint representation and gradient regularizations.

\subsection{Baseline Multilingual NMT Model}
\label{sec:baseline}
Following \citet{johnson2017google}, we concatenate all bilingual parallel corpora together to form the training set of a multilingual system, with an artificial token appended to each source sequence to specify the  target language.

Specifically, given a source sentence $\mathbf{x}^i=(x_1^i, x_2^i, ..., x_{|\mathbf{x}^i|}^i)$ in language $i$ and the parallel target sentence $\mathbf{y}^j=(y_1^j, y_2^j, ..., y_{|\mathbf{y}^j|}^j)$ in language $j$, the multilingual model is trained with the following cross-entropy loss:
\begin{equation}
    L_{\mathrm{NMT}} = -\sum_{t=1}^{|\mathbf{y}^j|} \log P_{\theta} (y_t^j ~|~ \mathbf{x^i}, \langle j \rangle, y_{1..(t-1)}^j),
    \label{eq:nmt}
\end{equation}
where $\langle j \rangle$ is the artificial token specifying the desired target language, and $P_{\theta}$ is parameterized using an encoder-decoder architecture based on a state-of-the-art Transformer backbone~\cite{vaswani2017attention}.

We then train the multilingual system on the concatenated parallel corpus of all available language pairs in both forward and backward directions, which is also referred to as a many-to-many multilingual system.
To balance the training batches between high-resource and low-resource language pairs, we adopt a temperature-based sampling to up/down-sample bilingual data accordingly~\cite{arivazhagan2019massively}. We set the temperature $\tau=5$ for all of our experiments. 

\subsection{Representation-Level Regularization: Target Language Prediction (TLP)}
As shown in Table \ref{tab:offtarget}, the current multilingual baseline faces serious off-target translation issue across the zero-shot directions.
With the multilingual decoder generating tokens in a wrong language, its decoder states for different target languages are also mixed and not well separated, in spite of the input token $\langle j \rangle$.
We thus introduce a representation-level regularizaion by adding an auxiliary Target Language Prediction (TLP) task to the standard NMT training.

Specifically, given the source sentence $\mathbf{x}=(x_1, x_2, ..., x_{|\mathbf{x}|})$ and a desired target language $\langle j \rangle$, the model generates a sequence of decoder states $\mathbf{z}=(z_1, z_2, ..., z_{|\mathbf{\hat{\mathbf{y}}}|})$\footnote{$\mathbf{z}$ is of the same length as the system translation $\hat{\mathbf{y}}$.}.
As the system feeds $\mathbf{z}$ through a classifier and predicts tokens (in Equation~\ref{eq:nmt}), we feed $\mathbf{z}$ through a LangID model to classify the desired target language $\langle j \rangle$.
TLP is then optimized with the cross-entropy loss:
\begin{equation}
    L_{\mathrm{TLP}} = - \log M_{\theta}(\mathbf{z}, \langle j \rangle),
\end{equation}
where the LangID model $M_{\theta}$ is parameterized as a 2-layer Transformer encoder with a LangID classifier on the top. TLP loss is then linearly combined with $L_{\mathrm{NMT}}$ with a coefficient $\alpha$.
Empirically, we found that any value from $\{0.1, 0.2, 0.3\}$ for $\alpha$ performs similarly well.
\begin{equation}
    L = (1-\alpha) \cdot L_{\mathrm{NMT}} + \alpha \cdot L_{\mathrm{TLP}}.
    \label{eq:loss}
\end{equation}

\paragraph{Implementation}
We implement the LangID model as a 2-layer Transformer encoder with input from the multilingual decoder states to classify the target language.
We add to the decoder states a sinusoidal positional embedding for position information.
We implement two common approaches to do classification: \emph{CLS\_Token} and \emph{Meanpooling}.
For CLS\_Token, we employ a BERT-like~\cite{devlin2018bert} CLS token and feed its topmost states to the classifier.
For Meanpooling, we simply take the mean of all output states and feed it to the classifer.
Their comparison is shown in Section~\ref{sec:tlp}.

\begin{algorithm}[t!]
    \SetAlgoLined
    \DontPrintSemicolon
    \SetKwInOut{Input}{Input}
    \SetCommentSty{itshape}
    \SetKwComment{Comment}{$\triangleright$ }{}
    \Input{Involved language set $\mathcal{L}$; Pre-trained model $\theta$; Training data $\mathcal{D}_{\text{train}}$; Oracle data $\mathcal{D}_{\text{oracle}}$; Update frequency $\mathit{n}$ }
    Initialize step $t = 0$, $\theta_0 = \theta$\\
    \While{not converged}{
        \Comment{Update oracle data gradients}
        \If{$t \pmod{\mathit{n}}= 0 $}{
            \For{$i$ in $\mathcal{L}$}{
                $g_{\text{oracle}}^i = \sum_{\mathcal{B}^i \sim \mathcal{D}_{\text{oracle}}^i} \nabla_{\theta_t} L(\theta_t, \mathcal{B}^i)$
            }
        }
        Sample minibatches grouped by target language $\mathcal{B} = \{ \mathcal{T}_i \}$ \\
        \For{$i$ in $\mathcal{L}$}{
            $g_{\text{train}}^i = \nabla_{\theta_t} L(\theta_t, \mathcal{T}^i)$ \\
            \If{$g_{\text{oracle}}^i \cdot g_{\text{train}}^i < 0$}{
                ${g}_{\text{train}}^i = g_{\text{train}}^i - \frac{g_{\text{train}}^i \cdot g_{\text{oracle}}^i}{\| g_{\text{oracle}}^i \|^2} ~ g_{\text{oracle}}^i$
            }
        }
        Update $t \leftarrow t + 1$ \\
        Update $\theta_t$ with gradient $\sum_i {g}_{\mathrm{train}}^i$ \\
    }
    \caption{Target-Gradient-Projection}
    \label{alg:tgp}
\end{algorithm}

\subsection{Gradient-Level Regularization: Target-Gradient-Projection (TGP)}
Although the TLP loss helps build more separable decoder states, it lacks reference signals to directly guide the system on how to model different target languages.
Inspired by recent gradient-alignment-based methods~\cite{wang2020balancing,yu2020gradient,wang2020gradient,wang2021gradient}, we propose Target-Gradient-Projection (TGP) to guide the model training with constructed oracle data, where we project the training gradient to not conflict with the oracle gradient.

\paragraph{Creation of oracle data}
Similar to \citet{wang2020balancing,wang2021gradient}, we build the oracle data from multilingual dev set, since the dev set is often available and is of a higher quality than the training set.
More importantly, for some zero-shot pairs, we are able to include hundreds or thousands of parallel samples from the dev set.
We construct the oracle data by concatenating all available dev sets and grouping them by the target language.
For example, the oracle data for French would include every other language to French.
The detailed construction of oracle data is specific to each dataset, and described in Section~\ref{sec:oracle}.
The dev set often serves to select the best checkpoint for training, thus we split the dev set as 80\% for oracle data and 20\% for checkpoint selection.

\paragraph{Implementation}
Contrary to standard multilingual training, where a training batch consists of parallel data from different language pairs, we group the training data by the \emph{target} language after the temperature-based sampling (Section~\ref{sec:baseline}).
By doing so, we treat the multilingual system as a multi-task learner, and translations into different languages are regarded as different tasks. Similarly, we construct the oracle data individually for each target language, whose gradients would serve as guidance to the training gradients.

For each step, we obtain the training gradients $g_{\mathrm{train}}^i$ for target language $i$, and the gradients of the corresponding oracle data $g_{\mathrm{oracle}}^i$.
Whenever we observe a conflict between $g_{\mathrm{train}}^i$ and $g_{\mathrm{oracle}}^i$, which is defined as a negative cosine similarity, we project $g_{\mathrm{train}}^i$ into the normal plane of $g_{\mathrm{oracle}}^i$ to de-conflict~\cite{yu2020gradient}.
\begin{equation}
    {g}_{\mathrm{train}}^i = g_{\mathrm{train}}^i - \frac{g_{\mathrm{train}}^i \cdot g_{\mathrm{oracle}}^i}{\| g_{\mathrm{oracle}}^i \|^2} ~ g_{\mathrm{oracle}}^i.
\end{equation}
The detailed algorithm is illustrated in Algorithm~\ref{alg:tgp}. We train the multilingual system with NMT loss or NMT+TLP joint loss for 40k steps before starting the TGP training, since the gradients of oracle data are not stable when trained from scratch. We set the update frequency $n=200$ for all our experiments. In our experiments, TGP is approximately 1.5x slower than the standard NMT training.


\section{Experimental Setup}

\subsection{Datasets: WMT-10}
\label{sec:wmt}
Following~\citet{wang2020multi}, we collect parallel data from publicly  available WMT campaigns\footnote{\url{http://www.statmt.org/wmt19/translation-task.html}} to form an English-centric multilingual WMT-10 dataset, including English and 10 other languages: French (Fr), Czech (Cs), German (De), Finnish (Fi), Latvian (Lv), Estonian (Et), Romanian (Ro), Hindi (Hi), Turkish (Tr) and Gujarati (Gu).
The size of bilingual datasets range from 0.08M to 10M, with five language pairs above 1M (Fr, Cs, De, Fi, Lv) and five language pairs below 1M (Et, Ro, Hi, Tr, Gu).
We use the same dev and test set as ~\citet{wang2020multi}.
Since the WMT data does not include zero-shot dev or test set, we have created 1k multi-way aligned dev and test sets for all involved languages based on the WMT2019 test set.
For evaluation, we picked 6 language pairs (12 translation directions) to examine zero-shot performance, including pairs of both high-resource languages (Fr-De and De-Cs), pairs of high- and low-resource languages (Ro-De and Et-Fr), and pairs of both low-resource languages (Et-Ro and Gu-Tr).
The detailed dataset statistics can be found in Section~\ref{appen:wmt}.

\begin{table}[t]
	\centering
	\begin{tabular}{cccc}
        \toprule
	    Code & \#Test & \#Dev & Overlap(\%) \\
        \midrule
        Tk & 1852 & 1852 & 97.46\% \\
        Ig & 1843 & 1843 & 96.31\% \\
        Li & 2000 & 2000 & 87.75\% \\
        Yi & 2000 & 2000 & 83.95\% \\
        Zu & 2000 & 2000 & 83.45\% \\
        \bottomrule
	\end{tabular}
	\caption{The top-5 most overlapped dev and test set on OPUS-100, the overlapping rate is calculated as the percentage of dev set that appears in the test. The average overlapping rate between dev and test is 15.26\% across all language pairs.}
	\label{tab:opus}
\end{table}

\subsection{Datasets: OPUS-100}
\label{sec:opus}
To evaluate our approaches in the massive multilingual settings, we adopt the OPUS-100 corpus from ~\citet{zhang2020improving}\footnote{\url{https://opus.nlpl.eu/opus-100.php}}.
OPUS-100 is also an English-centric dataset consisting of parallel data between English and 100 other languages.
We removed 5 languages (An, Dz, Hy, Mn, Yo) from OPUS, since they are not paired with a dev or test set.
However, while constructing the oracle data from its multilingual dev set, we found that the dev and test sets of OPUS-100 are noticeably noisy since they are directly sampled from web-crawled OPUS collections\footnote{\url{https://opus.nlpl.eu/}}.
As shown in Table~\ref{tab:opus}, several dev sets have  significant overlaps with their test sets. 15.26\% of dev set samples appear in the test set on average across all language pairs.
This is a significant flaw of the OPUS-100 (v1.0) that previous works have not noticed. To fix this, we rebuild the OPUS dataset as follows.
Without significantly modifying the dataset, we add an additional step of de-duplicating both the training and dev sets against the test\footnote{We keep the OPUS-100 test set as is to make a fair comparison against previous works, although there are also noticeable duplicates within the test set.}, and moving data from training set to complement the dev set due to de-duplication. We additionally sampled 2k zero-shot dev set using OPUS sampling scripts\footnote{\url{https://github.com/EdinburghNLP/opus-100-corpus}} to match the released 2k zero-shot test set.
The detailed dataset statistics can be found in section\footnote{Our rebuilt OPUS dataset is released at \url{https://github.com/yilinyang7/fairseq_multi_fix}.}.

\subsection{Training and Evaluation}
For both WMT-10 and OPUS-100, we tokenize the dataset with the SentencePiece model~\cite{kudo2018sentencepiece} and form a shared vocabulary of $64$k tokens.
We employ the Transformer-Big setting~\cite{vaswani2017attention} in all our experiments on the open-sourced Fairseq Implementation\footnote{\url{https://github.com/pytorch/fairseq}}~\cite{ott2019fairseq}.
The model is optimized using Adam~\cite{kingma2014adam} with a learning rate of $5 \times 10^{-4}$ and 4000 warm-up steps. The multilingual model is trained on 8 V100 GPUs with a batch size of 4096 tokens and a gradient accumulation of 16 steps, which effectively simulates the training on 128 V100 GPUs.
Our baseline model is trained with 50k steps, while it usually converges much earlier.
For evaluation, we employ beam search decoding with a beam size of 5 and a length penalty of 1.0. The BLEU score is measured by the de-tokenized case-sensitive SacreBLEU\footnote{BLEU+case.mixed+lang.{src}-{tgt}+numrefs.1+smooth.exp+tok.13a+version.1.4.14}~\cite{post2018call}.

In order to evaluate the off-target translations, we utilize  off-the-shelf LangID model from FastText~\cite{joulin2016fasttext} to detect the languages of translation outputs.

\subsection{Construction of Oracle Data}
\label{sec:oracle}
On WMT-10, we use our human labelled multi-way dev set together with the original English-centric WMT dev set to construct the oracle data.
On OPUS-100, we similarly combine the zero-shot dev set with original OPUS dev set for oracle data.
On OPUS, we further merge oracle data that consists of only English-centric dev sets, since it empirically obtains similar performance while exhibiting noticeable speedups.
The statistics of the constructed oracle data is shown in Section~\ref{appen:oracle}.

\section{Results}
In this section, we will demonstrate the effectiveness of our approach on both WMT-10 and OPUS-100 datasets.
The full results are documented separately in Tables~\ref{tab:wmt_enx},~\ref{tab:wmt_xen},
and \ref{tab:wmt_xy} for WMT-10, and Tables~\ref{tab:opus_en} and \ref{tab:opus_xy} for OPUS-100.

\begin{table}
    \centering
    \small
    \begin{tabular}{lcc}
    \toprule
    \textbf{TLP} & $\alpha$ & Avg. BLEU \\
    \midrule
    Baseline & 0 & 23.58 \\
    \midrule
    \multirow{3}{*}{Meanpooling} & 0.1 & 23.85\\
    & 0.2 & 23.73 \\
    & 0.3 & \textbf{23.93}\\
    \midrule
    \multirow{3}{*}{CLS\_Token} & 0.1 & 23.81 \\
    & 0.2 & 23.76 \\
    & 0.3 & \textit{Diverged} \\
    \bottomrule
    \end{tabular}
    \caption{Comparing TLP approaches on WMT-10. BLEU is averaged across all English-centric directions.}
    \label{tab:tlp}
\end{table}

\begin{table}
    \centering
    \small
    \begin{tabular}{lc}
    \toprule
    \textbf{TGP} &  Avg. BLEU \\
    \midrule
    Baseline & 23.58 \\
    \midrule
    Model-wise & \textbf{24.35}\\
    Layer-wise & 23.77\\
    Matrix-wise & 23.90\\
    \bottomrule
    \end{tabular}
    \caption{Comparing TGP granularity on WMT-10. BLEU is averaged across all English-centric directions.}
    \label{tab:tgp}
\end{table}

\begin{table*}[h]
    \centering
    \setlength{\tabcolsep}{5pt}
    \small
    \begin{tabular}{cl cc cccccccccc ss}
    \toprule
    & \textbf{En $\to$ X }& {\small TLP} & {\small TGP} & \small Fr &\small  Cs &\small  De &\small  Fi &\small  Lv &\small  Et &\small  Ro &\small  Hi &\small  Tr &\small  Gu &\small  Avg & \small Off-Tgts \\
    \midrule
    {\scriptsize\texttt{1}} & Bilingual & - & -   & 36.3  & 22.3  & 40.2  & 15.2  & 16.5  & 15.0  & 23.0  & 12.2  & 13.3  & 7.9 & 20.19 & 1.00\% \\
    \cmidrule{2-16}
    {\scriptsize\texttt{2}} & Baseline & - & - & 32.7 & 19.8 & 38.4 & 14.2 & 17.4 & 19.9 & 25.5 & 13.7 & 16.3 & 11.6 & 20.95 & 1.06\% \\
    {\scriptsize\texttt{3}} & + Finetune & - & - & 21.2 & 13.0 & 24.4 & 10.4 & 12.1 & 14.6 & 23.4 & 11.6 & 10.5 & 17.4 & 15.86 & 0.86\% \\
    \cmidrule{2-16}
    {\scriptsize\texttt{4}} & \multirow{5}{*}{\shortstack[l]{Ours\\ {\scriptsize ( Baseline + \_\_\_ )}}} & \checkmark & - & 33.2 & 20.1 & 38.9 & 14.7 & 17.7 & 20.1 & 26.1 & 13.5 & 16.5 & 12.7 & 21.35 & 1.09\%  \\
    {\scriptsize\texttt{5}} & & - & \checkmark & 33.6 & 20.0 & 38.7 & 14.7 & 17.6 & 20.3 & 25.8 & 16.0 & 16.5 & 18.6 & \textbf{22.18} & 0.94\% \\
    {\scriptsize\texttt{6}} & & \checkmark & \checkmark & 33.1 & 20.0 & 38.7 & 15.2 & 17.5 & 20.4 & 26.4 & 14.8 & 16.3 & 18.5 & 22.09 & 0.98\% \\
    \cmidrule{3-16}
    {\scriptsize\texttt{7}} & & - & \checkmark$^{\star}$ & 32.8 & 20.1 & 37.4 & 14.8 & 17.7 & 19.7 & 25.8 & 15.7 & 16.4 & 18.4 & 21.88 & 0.92\% \\
    {\scriptsize\texttt{8}} & & \checkmark & \checkmark$^{\star}$ & 33.0 & 20.2 & 37.8 & 15.1 & 17.7 & 20.2 & 26.3 & 14.6 & 16.5 & 19.4 & \textbf{22.08} & 0.97\% \\
    \bottomrule
    \end{tabular}
    \caption{BLEU scores of English $\to$ 10 languages translation on WMT-10. \checkmark$^{\star}$ denotes TGP training in a zero-shot manner for all evaluated English-free pairs. ``Off-Tgts'' column reports the average off-target rates from FastText LangID model, while the off-target rate on the references is 0.81\%. }
    \label{tab:wmt_enx}
\end{table*}

\begin{table*}[h]
    \centering
    \setlength{\tabcolsep}{5pt}
    \small
    \begin{tabular}{cl cc cccccccccc ss}
    \toprule
    & \textbf{X $\to$ En}& {\small TLP} & {\small TGP} & {\small Fr} & {\small Cs} & \small De & \small Fi & \small Lv & \small Et & \small Ro & \small Hi & \small Tr & \small Gu & \small Avg & \small Off-Tgts \\
    \midrule
    {\scriptsize\texttt{1}} & Bilingual & - & - & 36.2  & 28.5  & 40.2  & 19.2  & 17.5  & 19.7  & 29.8  & 14.1  & 15.1  & 9.3 & 22.96 & 0.30\% \\
    \cmidrule{2-16}
    {\scriptsize\texttt{2}} & Baseline & - & - & 34.0 & 28.2 & 39.1 & 19.9 & 19.5 & 24.8 & 34.6 & 21.9 & 22.4 & 17.8 & 26.22 & 0.23\% \\
    {\scriptsize\texttt{3}} & + Finetune & - & - & 24.7 & 22.3 & 30.1 & 16.9 & 16.2 & 21.1 & 39.4 & 17.7 & 17.6 & 17.0 & 22.30 & 0.13\% \\
    \cmidrule{2-16}
    {\scriptsize\texttt{4}} & \multirow{5}{*}{\shortstack[l]{Ours\\ {\scriptsize ( Baseline + \_\_\_ )}}} & \checkmark & - & 35.0 & 28.7 & 39.5 & 20.4 & 20.2 & 25.5 & 34.6 & 21.4 & 22.3 & 17.4 & 26.50 & 0.20\% \\
    {\scriptsize\texttt{5}} & & - & \checkmark & 34.5 & 29.1 & 40.0 & 20.8 & 20.1 & 26.3 & 39.5 & 23.4 & 22.8 & 19.5 & 27.60 & 0.20\% \\
    {\scriptsize\texttt{6}} & & \checkmark & \checkmark & 34.2 & 29.4 & 39.5 & 21.3 & 20.3 & 26.0 & 40.4 & 24.1 & 23.0 & 19.8 & \textbf{27.80} & 0.19\% \\
    \cmidrule{3-16}
    {\scriptsize\texttt{7}} & & - & \checkmark$^{\star}$ & 33.9 & 28.7 & 38.8 & 21.0 & 20.0 & 26.4 & 39.5 & 24.0 & 22.6 & 19.2 & 27.41 & 0.15\% \\
    {\scriptsize\texttt{8}} & & \checkmark & \checkmark$^{\star}$ & 34.4 & 28.8 & 39.6 & 21.3 & 20.5 & 26.8 & 40.6 & 24.2 & 22.9 & 20.8 & \textbf{27.99} & 0.20\% \\
    \bottomrule
    \end{tabular}
    \caption{BLEU scores of 10 languages $\to$ English translation on WMT-10. \checkmark$^{\star}$ denotes TGP training in a zero-shot manner for all evaluated English-free pairs. ``Off-Tgts'' column reports the average off-target rates from FastText LangID model, while the off-target rate on the references is 0.12\%.}
    \label{tab:wmt_xen}
\end{table*}

\begin{table*}[h]
    \centering
    \setlength{\tabcolsep}{4pt}
    \resizebox{\textwidth}{!}{
    \begin{tabular}{cl cc cccccccccccc ss}
    \toprule
    & \multirow{2}{*}{\textbf{En-Free}} & \small{\multirow{2}{*}{TLP}} & \small{\multirow{2}{*}{TGP}} & \multicolumn{2}{c}{\small Fr-De} & \multicolumn{2}{c}{\small De-Cs} & \multicolumn{2}{c}{\small Ro-De} & \multicolumn{2}{c}{\small Et-Fr}  & \multicolumn{2}{c}{\small Et-Ro} & \multicolumn{2}{c}{\small Gu-Tr} & \small BLEU & \small Off-Tgt \\
    & & & & \small $\gets$ & \small $\to$ & \small $\gets$ & \small $\to$ & \small $\gets$ & \small $\to$ & \small $\gets$ & \small $\to$ & \small $\gets$ & \small $\to$ & \small $\gets$ & \small $\to$ & \small Avg & \small Avg (\%) \\
    \midrule
    {\scriptsize\texttt{1}} & Pivoting  & - & -  & 24.9 & 19.3 & 19.4 & 18.9 & 19.1 & 18.8 & 16.2 & 20.9 & 16.4 & 16.8 & 5.2 & 6.4 & 16.86 & 1.1\%\\
    \cmidrule{2-18}
    {\scriptsize\texttt{2}} & Baseline & - & -  & 18.5 & 12.8 & 15.8 & 13.6 & 17.5 & 16.0 & 10.3 & 13.7 & 12.5 & 14.4 & 0.9 & 1.9 & 12.33 & 24.5\% \\
    {\scriptsize\texttt{3}} & + Finetune & - & -  & 17.9 & 15.1 & 14.6 & 13.1 & 18.7 & 14.6 & 12.3 & 16.4 & 12.6 & 17.0 & 8.1 & 7.0 & 13.95 & \textbf{0.7\%} \\
    \cmidrule{2-18}
    {\scriptsize\texttt{4}} & \multirow{5}{*}{\shortstack[l]{Ours\\ {\scriptsize ( Baseline + \_ )}}} & \checkmark & - & 21.4 & 15.7 & 17.8 & 15.8 & 18.1 & 17.0 & 14.1 & 17.3 & 14.0 & 15.5 & 3.0 & 3.8 & 14.45 & 6.0\% \\
    {\scriptsize\texttt{5}} & & - & \checkmark & 25.4 & 19.4 & 20.0 & 18.7 & 22.7 & 19.6 & 16.5 & 22.4 & 17.0 & 20.2 & 6.2 & 6.7 & 17.90 & 0.9\% \\
    {\scriptsize\texttt{6}} & & \checkmark & \checkmark & 25.2 & 19.7 & 19.7 & 18.9 & 23.0 & 19.8 & 16.1 & 21.4 & 16.7 & 20.8 & 6.4 & 7.3 & \textbf{17.92} & 0.9\%\\
    \cmidrule{3-18}
    {\scriptsize\texttt{7}} & & - & \checkmark$^{\star}$ & 24.5 & 18.5 & 18.8 & 18.2 & 21.8 & 18.2 & 16.6 & 20.9 & 16.1 & 19.5 & 5.6 & 6.9 & 17.13 & 2.0\%  \\
    {\scriptsize\texttt{8}} & & \checkmark & \checkmark$^{\star}$ & 24.2 & 18.3 & 20.0 & 18.3 & 22.3 & 19.5 & 15.9 & 21.5 & 16.1 & 20.1 & 5.6 & 7.1 & \textbf{17.41} & 2.1\% \\
    \bottomrule
    \end{tabular}}
    \caption{BLEU scores of English-free translations on WMT-10. \checkmark$^{\star}$ denotes TGP training in a zero-shot manner for all evaluated English-free pairs. As a reference, the average off-target rate reported by FastText LangID model is 0.68\% on the references.}
    \label{tab:wmt_xy}
\end{table*}

\begin{table*}[h]
    \centering
    \small
    \begin{tabular}{cl cc cccs|cccs}
    \toprule
    & \multirow{2}{*}{\textbf{English-Centric}} & \multirow{2}{*}{TLP} & \multirow{2}{*}{TGP} & \multicolumn{4}{c}{X $\to$ English} & \multicolumn{4}{c}{English $\to$ X}  \\
    \cmidrule{5-12}
    & & & & High & Med & Low & All & High & Med & Low & All \\
    \midrule
    {\scriptsize\texttt{1}} & \citet{zhang2020improving} (24L)  & - & - & 30.29  & 32.58  & 31.90  & 31.36  & 23.69  & 25.61  & 22.24 & 23.96 \\
    \midrule
    {\scriptsize\texttt{2}} & Baseline & - & -  & 30.27 & 33.50 & 31.94 & 31.61 & 23.64 & 29.13 & 29.38 & 26.56 \\
    {\scriptsize\texttt{3}} & + Finetune & - & - & 19.85 & 29.93 & 36.49 & 26.57 & 15.45 & 23.84 & 30.05 & 21.21 \\
    \cmidrule{2-12}
    {\scriptsize\texttt{4}} & \multirow{5}{*}{\shortstack[l]{Ours\\ {\scriptsize ( Baseline + \_\_ )}}} & \checkmark & - & 30.31 & 33.17 & 33.06 & 31.78 & 23.71 & 29.11 & 29.24 & 26.55 \\
    {\scriptsize\texttt{5}} & & - & \checkmark & 30.17 & 38.21 & 42.23 & 35.26 & 23.51 & 30.61 & 33.60 & 27.88 \\
    {\scriptsize\texttt{6}} & & \checkmark & \checkmark & 29.96 & 38.32 & 41.94 & 35.13 & 23.35 & 30.23 & 33.63 & 27.69 \\
    \cmidrule{3-12}
    {\scriptsize\texttt{7}} & & - & \checkmark$^{\star}$ & 30.07 & 38.28 & 42.25 & 35.24 & 23.53 & 30.53 & 33.97 & 27.95 \\
    {\scriptsize\texttt{8}} & & \checkmark & \checkmark$^{\star}$ & 29.83 & 38.50 & 42.51 & 35.24 & 23.30 & 30.28 & 33.43 & 27.64 \\
    \bottomrule
    \end{tabular}
    \caption{Average test BLEU for High/Medium/Low-resource language pairs on OPUS-100 dataset. \emph{All} denotes the average BLEU for all langugage pairs. \checkmark$^{\star}$ denotes TGP training in a zero-shot manner for all evaluated English-free pairs.}
    \label{tab:opus_en}
\end{table*}

\begin{table*}[h]
    \centering
    \setlength{\tabcolsep}{4pt}
    \small
    \begin{tabular}{cl cc cccccccccccc ss}
    \toprule
    & \multirow{2}{*}{\textbf{En-Free}} & \small{\multirow{2}{*}{\small TLP}} & \small{\multirow{2}{*}{\small TGP}} & \multicolumn{2}{c}{\small De-Fr} & \multicolumn{2}{c}{\small Ru-Fr} & \multicolumn{2}{c}{\small Nl-De} & \multicolumn{2}{c}{\small Zh-Ru}  & \multicolumn{2}{c}{\small Zh-Ar} & \multicolumn{2}{c}{\small Nl-Ar} & \small BLEU & \small Off-Tgt \\
    & & & & \small $\gets$ & \small $\to$ & \small $\gets$ & \small $\to$ & \small $\gets$ & \small $\to$ & \small $\gets$ & \small $\to$ & \small $\gets$ & \small $\to$ & \small $\gets$ & \small $\to$ & \small Avg & \small Avg (\%) \\
    \midrule
    {\scriptsize\texttt{1}} & Pivoting  & - & -  & 18.5 & 21.5 & 21.0 & 26.7 & 21.7 & 19.7 & 13.6 & 20.2 & 14.9 & 17.8 & 16.6 & 5.7 & 18.16 & 4.2\% \\
    \midrule
    {\scriptsize\texttt{2}} & Baseline & - & -  & 3.5 & 2.9 & 7.2 & 4.0 & 4.9 & 4.8 & 4.5 & 11.0 & 2.8 & 11.1 & 1.5 & 2.6 & 5.07 &  65.8\% \\
    {\scriptsize\texttt{3}} & + Finetune & - & - & 13.1 & 14.9 & 15.9 & 17.3 & 16.0 & 14.4 & 12.8 & 14.8 & 15.2 & 12.9 & 8.5 & 3.1 & 13.24 & \textbf{3.7\%} \\
    \cmidrule{2-18}
    {\scriptsize\texttt{4}} & \multirow{5}{*}{\shortstack[l]{Ours\\ {\scriptsize ( Baseline + \_\_ )}}} & \checkmark & - & 3.9 & 3.3 & 8.0 & 6.4 & 6.1 & 5.0 & 7.7 & 10.6 & 3.2 & 11.2 & 1.3 & 3.4 & 5.84 & 60.5\% \\
    {\scriptsize\texttt{5}} & & - & \checkmark & 16.1 & 18.2 & 18.6 & 21.1 & 19.1 & 18.5 & 13.9 & 16.2 & 15.0 & 14.4 & 12.3 & 5.0 & \textbf{15.70} & 4.8\% \\
    {\scriptsize\texttt{6}} & & \checkmark & \checkmark & 16.4 & 17.7 & 18.3 & 21.0 & 19.0 & 18.3 & 12.7 & 16.2 & 14.2 & 14.4 & 12.2 & 5.0 & 15.45 & 4.7\% \\
    \cmidrule{3-18}
    {\scriptsize\texttt{7}} & & - & \checkmark$^{\star}$ & 4.2 & 12.2 & 12.8 & 20.9 & 10.0 & 5.9 & 11.5 & 13.5 & 12.8 & 12.6 & 11.8 & 4.5 & 11.06 & 31.1\% \\
    {\scriptsize\texttt{8}} & & \checkmark & \checkmark$^{\star}$ & 6.6 & 14.2 & 16.7 & 21.4 & 16.2 & 8.6 & 12.9 & 14.2 & 14.7 & 12.6 & 11.8 & 4.6 & \textbf{12.88} & \textbf{16.9\%} \\
    \bottomrule
    \end{tabular}
    \caption{BLEU scores of English-free translations on OPUS-100. \checkmark$^{\star}$ denotes TGP training in a zero-shot manner for all evaluated English-free pairs. As a reference, the average off-target rate reported by FastText LangID model is 4.85\% on the references.}
    \label{tab:opus_xy}
\end{table*}

\subsection{TLP Results}
\label{sec:tlp}
\paragraph{Hyper-parameter Tuning}Table~\ref{tab:tlp} shows the comparison between TLP implementations on the WMT-10 dev set.
We observe that the Meanpooling approach for TLP is both more stable and delivers slightly better performance.
In all the following experiments, we use the Meanpooling approach for TLP, with $\alpha=0.3$ on WMT-10 and $\alpha=0.1$ on OPUS-100.

\paragraph{Performance}
From Tables~\ref{tab:wmt_enx} and \ref{tab:wmt_xen} (row 4 vs. row 2), we can see that TLP outperforms baselines in most En-X and X-En directions and all English-free directions as shown in Table~\ref{tab:wmt_xy} (row 4 vs. row 2).
On average, TLP gains +0.4 BLEU on En-X, +0.28 BLEU on X-En and +2.12 BLEU on English-free directions.
TLP also significantly reduces the off-target rate from 24.5\% down to 6.0\% (in Table~\ref{tab:wmt_xy}).
Meanwhile on OPUS-100, TLP performs similarly in English-centric directions (in Table~\ref{tab:opus_en}) while yielding +0.77 BLEU improvement on English-free directions, together with a 65.8\% $\to$ 60.5\% drop in off-target occurrences (in Table~\ref{tab:opus_xy}).

These results demonstrate that by adding an auxiliary TLP loss, multilingual models much better retain information about the target language, and moderately improved on English-free pairs.

\begin{figure}
    \vspace{-0.6cm}
	\centering
	\includegraphics[width=0.5\textwidth]{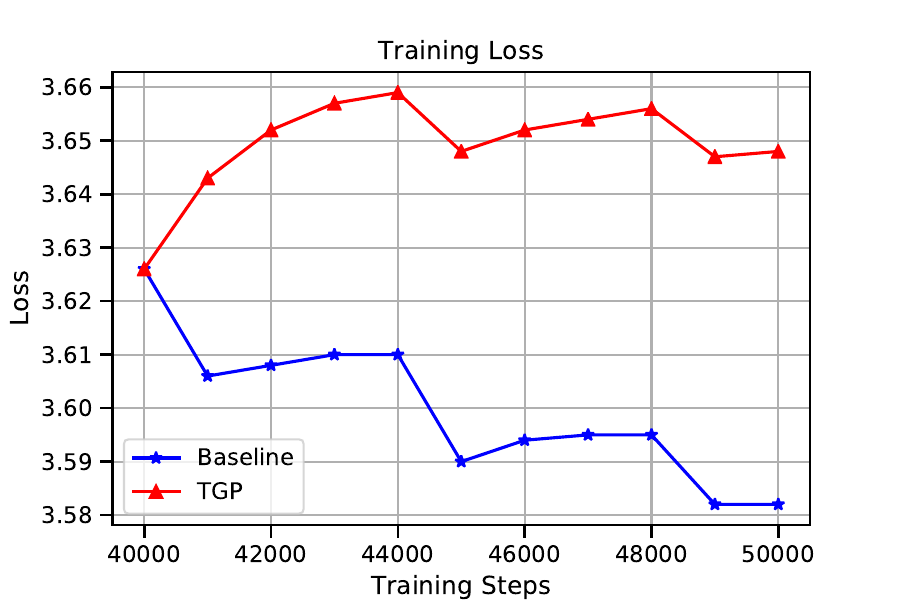}
	\caption{The training loss curve of TGP on WMT-10.}
	\label{fig:tgp_loss}
\end{figure}

\begin{figure}
	\centering
	\includegraphics[width=0.5\textwidth]{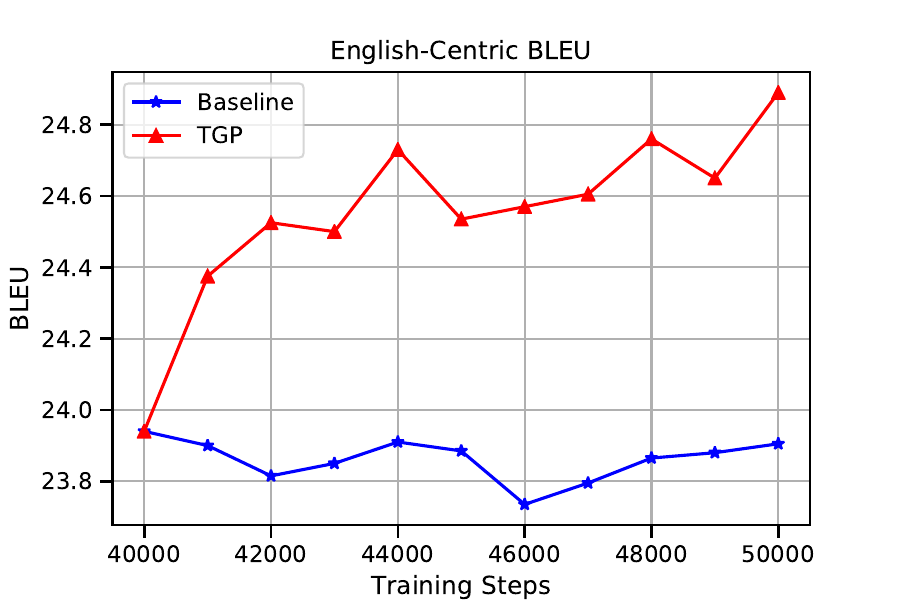}
	\includegraphics[width=0.5\textwidth]{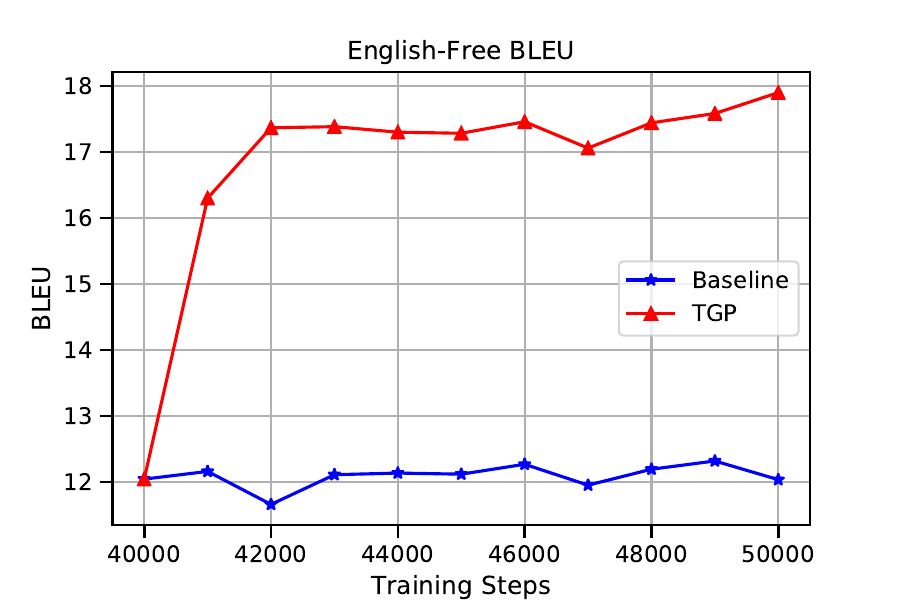}
	\caption{The test BLEU curves of TGP on WMT-10.}
	\label{fig:tgp_bleu}
\end{figure}

\subsection{TGP Results}
\paragraph{Settings}
Similar to \citet{yu2020gradient,wang2020gradient}, the conflict detection and de-conflict projection of TGP training could be done with different granularities.
We compare three options: (1) model-wise: flatten all parameters into one vector, and perform projection on the entire model; (2) layer-wise: perform individually for each layer of encoder and decoder; (3) matrix-wise: perform individually for each parameter matrix.
From Table~\ref{tab:tgp}, we found operating on the model-level gives the best performance, and as a result all our TGP experiments are done on the model-level.
We then perform TGP training for 10k steps on the 40k-step pretrained model.

\paragraph{Performance}
In Tables~\ref{tab:wmt_enx}, \ref{tab:wmt_xen} and \ref{tab:wmt_xy} (row 5 vs. 2), TGP gains significant improvements on all directions of WMT-10: averaging +1.23 BLEU on En-X, +1.38 BLEU on X-En and +5.57 BLEU on English-free directions, while also reducing the off-target rates from 24.5\% down to only 0.9\%.
Similar gains could also be found on OPUS-100 (in Tables~\ref{tab:opus_en} and \ref{tab:opus_xy}): +3.65 BLEU on En-X, +1.32 BLEU on X-En and +10.63 BLEU on English-free, and a whopping 65.8\% $\to$ 4.8\% reduction to off-target occurrences.
These results demonstrate the overwhelming effectiveness of TGP on all translation tasks as well as on reducing off-target cases.

\paragraph{Learning curves}
Figures~\ref{fig:tgp_loss} and \ref{fig:tgp_bleu} illustrate the learning curves of TGP on WMT-10.
Different from the baseline curves, TGP observes a slight \emph{increase} in training loss, which shows TGP as a regularizer prevents the model from overfitting to the training set.
Meanwhile, a steady increase on both English-centric and English-free test BLEU demonstrates the effectiveness of TGP training.

\paragraph{Finetuning on oracle data}
Suggested by \citet{wang2021gradient}, we explore another baseline usage of oracle data: direct finetuning.
For finetuning, we concatenate all oracle data from different target languages together.
With the same settings as TGP (finetuning for 10k steps on the 40k-step baseline), we also observe a noticeable improvement on English-free directions: an average of +1.62 BLEU on WMT-10 and +8.17 BLEU on OPUS-100, with the most reduction on the off-target occurrences (row 3 of Tables~\ref{tab:wmt_xy} and \ref{tab:opus_xy}).
However in comparison to TGP, directly finetuning on oracle data lacks the step of separately modeling for different target languages and the crucial step of de-conflicting, thus it hurts the English-centric (En-X and X-En) directions while also lagging as much as -3.95 BLEU on English-free pairs (Table~\ref{tab:wmt_xy}, row 3 vs. 5).

\subsection{TGP in a Zero-Shot Setting}
\label{sec:tgp_zeroshot}
Although TGP and Finetuning obtain significant reductions on off-target cases, they both assume some amount of direct parallel data on English-free pairs, while in reality, such direct parallel data may not exist in extreme low-resource scenario.
To simulate a zero-shot setting, we build a new oracle dataset that explicitly excludes parallel data of all evaluated English-free pairs.\footnote{We exclude the parallel data of 6 evaluation pairs from oracle data while keeping the others. More details in Section~\ref{appen:oracle}.}
In this setting, all evaluation pairs are trained in a strict zero-shot manner to test the system's generalization ability.

\paragraph{Performance}
In Tables~\ref{tab:wmt_enx}, \ref{tab:wmt_xen}, \ref{tab:wmt_xy} row 7 with \checkmark$^{\star}$, TGP in a zero-shot manner slightly lags behind TGP with full oracle data, while still gaining  significant improvement compared to the  baseline.
On average, we observe a gain of +0.93 BLEU on En-X, +1.19 BLEU on X-En and +4.8 on English-free compared to baseline (row 7 vs. 2), and a slight decrease of -0.3 BLEU on En-X, -0.19 BLEU on X-En and -0.77 BLEU on English-free compared to TGP with full oracle set (row 7 vs. 5).
Meanwhile on OPUS-100 (in Tables~\ref{tab:opus_en},\ref{tab:opus_xy}), we also observe a consistent gain against the baseline (row 7 vs. 2), but a noticeable -4.64 BLEU drop on English-free pairs against TGP with full oracle data (row 7 vs. 5).
The performance drop (zero-shot vs. full data) illustrates that thousands of parallel  samples\footnote{In our case, we obtain 1k for WMT and 2k for OPUS.} could greatly help TGP on zero-shot translations, and we suspect the drop of only -0.77 BLEU on WMT-10 is due to the multi-way nature of our WMT oracle data.
Meanwhile, TGP in a zero-shot setting is still shown to greatly improve translation performance and significantly reduces off-target occurrences (24.5\% $\to$ 2.0\% on WMT and 65.8\% $\to$ 31.1\% on OPUS).

\subsection{Joint TLP+TGP}
TLP models could be seamlessly adopted in TGP training, by replacing the original NMT loss with a joint NMT+TLP loss.
Comparing the joint TLP+TGP approach to TGP-only (row 6 vs. 5 in Tables~\ref{tab:wmt_enx}-
\ref{tab:opus_xy}), we observe no significant differences in the full oracle data scenario (changes within $\pm$ 0.3 BLEU).
However in zero-shot setting, the joint TLP+TGP approach noticeably outperforms TGP-only by +1.82 BLEU on average in English-free pairs (Table~\ref{tab:opus_xy},  row 8 vs. 7).
Given TLP alone is only able to gain +0.77 BLEU (row 4 vs. 2), it hints TLP and TGP to have a \emph{synergy} effect in the extremely low resource scenario.

\section{Discussions on Off-Target Translations}
In this section, we will discuss the off-target translation in the English-centric directions and its relationship with token-level off-targets.

\subsection{Off-Targets on English-Centric Pairs}
Previous literature only studies the off-target translations in the zero-shot directions.
However, we show in Tables~\ref{tab:wmt_enx} and \ref{tab:wmt_xen} that off-target translation also occurs in the English-centric directions (although to a smaller scale).
Since we are using an imperfect LangID model, we quantify its error margin as the off-target rates reported on the references\footnote{We assume the WMT references are always in-target, since they are collected from human translators.}.
We could then observe that the baseline model is producing 0.25\% and 0.18\% more off-target translations than the references in En-X and X-En directions respectively, which are also reduced by our proposed TLP and TGP approaches.

\begin{table}
    \centering
    \small
    \begin{tabular}{lccc}
    \toprule
    \bf Token Off-Tgt & \bf En$\to$X & \bf X$\to$En & \bf En-Free \\
    \midrule
    Reference & 0.10\% & 0.02\% & 0.73\% \\
    \midrule
    Baseline & 0.08\% & 0.02\% & 0.16\% \\
    +TLP & 0.08\% & 0.02\% & 0.09\% \\
    +TGP & 0.08\% & 0.02\% & 0.07\% \\
    \bottomrule
    \end{tabular}
    \caption{Token-level off-target rates on WMT-10 quantified by whether appearing in the training set.}
    \label{tab:token_level}
\end{table}

\subsection{Token-Level Off-Targets}
\begin{figure}
    \vspace{-0.6cm}
	\centering
	\includegraphics[width=0.5\textwidth]{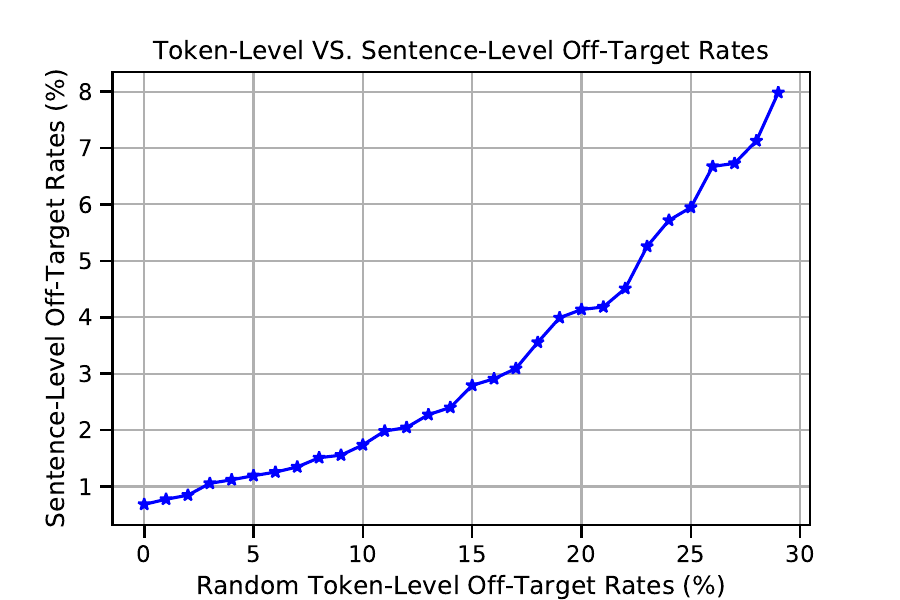}
	\caption{Relationship between the random token-level probability $p$ and the reported sentence-level off-target rates. Token-level off-targets are introduced by replacing the in-target token to a random off-target token with a probability $p$. Analysis done on the WMT-10 English-free references.}
	\label{fig:sensitive}
\end{figure}

Given a sentence-level LangID model, we are also curious about how it represents errors at the token level.
We attempt to simply quantify the token-level off-target rates by checking whether each token appears in the training data. Surprisingly,
results in  Table~\ref{tab:token_level} show that all systems contain lower token-level off-targets than the references. 
We hypothesize that it is attributed to two main reasons: 1) training set also contains noisy off-target tokens. 2) there are domain/vocabulary mismatches between training and test set, especially for the English-free pairs.

In order to test the robustness of the FastText LangID model as well as to relate our reported sentence-level scores to the token-level, we randomly introduced off-target tokens to the references and observed the sentence-level scores.
Specifically, we replaced the in-target token to a random off-target one with a probability $p$.
Figure~\ref{fig:sensitive} shows a near exponential curve between the sentence-level scores and probability $p$.
We could also observe that the sentence-level LangID model is somewhat robust to token-level off-target noises, e.g. it reports around 4\% off-target rates given 20\% of the tokens are replaced with off-target ones.

\section{Related Work}
\paragraph{Multilingual NMT}
Multilingual NMT aims to train one model to serve all language pairs~\cite{ha2016toward,firat2016multi,johnson2017google}.
Several subsequent works explored various parameter sharing strategies to mitigate the representation bottleneck~\cite{blackwood2018multilingual,platanios2018contextual,sachan2018parameter,sen2019multilingual}.
Meanwhile, there are also notorious cases of off-target translation especially in English-Free pairs.
Previous works either resort to back-translation techniques to generate synthetic English-Free data~\cite{gu2019improved,zhang2020improving}, or to model-level changes to the encoder-decoder alignments~\cite{arivazhagan2019massively,liu2020improving}.
In contrast, we propose a joint representation and gradient regularization approach to reduce off-target translations and significantly improve performance across all language pairs.

\paragraph{Multi-Task Learning for NMT}
Multi-task learning (MTL) is a widely used technique to share model parameters and improve generalization~\cite{ruder2017overview}.
For NMT, previous works have leveraged MTL to inject linguistic knowledge or leveraged monolingual data~\cite{eriguchi2017learning,niehues2017exploiting,kiperwasser2018scheduled,wang2020multi}.
Our work leverages an auxiliary TLP loss to help learn more separable model states for different target languages.

\paragraph{Optimization Learning}
Previous works have studied the optimization challenges in multi-task training~\cite{hessel2019multi,schaul2019ray}, where \citet{yu2020gradient} proposed to resolve gradient conflicts between different tasks.
Meanwhile for NMT, ~\citet{wang2021gradient,wang2020balancing} proposed to mask out or assign different weights to training samples based on the gradient alignments with validation set.
\citet{yu2020gradient} proposed to resolve the pair-wise gradient conflicts between translation directions.
In contrast, we propose TGP to guide the training process by projecting training gradients according to the oracle gradients.

\section{Conclusions}
In this work, we aimed to reduce the off-target translations with our proposed joint representation (TLP) and gradient (TGP) regularization to guide the internal modeling of target languages.
Our results showed both approaches to be highly effective at improving translation quality and to a large extent reduced the off-target occurrences.
As a future direction, we will investigate off-target translations during decoding time~\cite{yang2018breaking,yang2020sub}.

\section*{Acknowledgements}
The authors thank the anonymous reviewers for their insightful and helpful comments.

\bibliography{anthology,custom}
\bibliographystyle{acl_natbib}

\clearpage
\appendix
\section{Appendix}
\subsection{WMT-10 Data}
\label{appen:wmt}
We concatenate all resources except WikiTitles from WMT2019. For Fr and Cs, we randomly sample $10$M sentence pairs from the full corpus. The detailed statistics of bitext data can be found in Table~\ref{tab:wmt10}. We randomly sample $1,000$ sentence pairs from each individual validation set and concatenate them to construct a multilingual validation set.

\begin{table*}[h]
    \centering
    \begin{tabular}{ccccc}
    \toprule
    Code & Language & \#Train   & Dev  & Test \\
    \midrule
    Fr   & French   & $10$M      & Newstest13 & Newstest15 \\
    Cs   & Czech    & $10$M      & Newstest16 & Newstest18 \\
    De   & German   & $4.6$M     & Newstest16 & Newstest18 \\
    Fi   & Finnish  & $4.8$M     & Newstest16 & Newstest18 \\
    Lv   & Latvian  & $1.4$M     & Newsdev17 & Newstest17 \\
    Et   & Estonian & $0.7$M     & Newsdev18 & Newstest18 \\
    Ro   & Romanian & $0.5$M     & Newsdev16 & Newstest16 \\
    Hi   & Hindi    & $0.26$M    & Newsdev14 & Newstest14 \\
    Tr   & Turkish  & $0.18$M    & Newstest16 & Newstest18 \\
    Gu   & Gujarati & $0.08$M    & Newsdev19 & Newstest19 \\
    \bottomrule
    \end{tabular}
    \caption{Statistics of the WMT-10 dataset. \label{tab:wmt10}}
\end{table*}

\subsection{OPUS-100 Dataset}
\label{appen:opus}
After de-duplicating both training and dev set against the test set, the statistics of OPUS-100 dataset is shown in Table~\ref{tab:opus100}.

\begin{table*}[t]
    \centering
    \small
    \begin{tabular}{llrrrp{1cm}llrrr}
    \toprule
    Code & Language & Train & Dev & Test &  & Code & Language & Train & Dev & Test \\
    af & Afrikaans & 275451 & 2000 & 2000 & & lv & Latvian & 999976 & 2000 & 2000 \\
    am & Amharic & 88979 & 2000 & 2000 & & mg & Malagasy & 590759 & 2000 & 2000 \\
    ar & Arabic & 999988 & 2000 & 2000 & & mk & Macedonian & 999955 & 2000 & 2000 \\
    as & Assamese & 138277 & 2000 & 2000 & & ml & Malayalam & 822709 & 2000 & 2000 \\
    az & Azerbaijani & 262060 & 2000 & 2000 & & mr & Marathi & 26904 & 2000 & 2000 \\
    be & Belarusian & 67183 & 2000 & 2000 & & ms & Malay & 999973 & 2000 & 2000 \\
    bg & Bulgarian & 999983 & 2000 & 2000 & & mt & Maltese & 999941 & 2000 & 2000 \\
    bn & Bengali & 999990 & 2000 & 2000 & & my & Burmese & 23202 & 2000 & 2000 \\
    br & Breton & 153283 & 2000 & 2000 & & nb & Norwegian Bokmål & 142658 & 2000 & 2000 \\
    bs & Bosnian & 999991 & 2000 & 2000 & & ne & Nepali & 405723 & 2000 & 2000 \\
    ca & Catalan & 999983 & 2000 & 2000 & & nl & Dutch & 999984 & 2000 & 2000 \\
    cs & Czech & 999988 & 2000 & 2000 & & nn & Norwegian Nynorsk & 485547 & 2000 & 2000 \\
    cy & Welsh & 289007 & 2000 & 2000 & & no & Norwegian & 999977 & 2000 & 2000 \\
    da & Danish & 999991 & 2000 & 2000 & & oc & Occitan & 34886 & 2000 & 2000 \\
    de & German & 999993 & 2000 & 2000 & & or & Oriya & 14027 & 1317 & 1318 \\
    el & Greek & 999982 & 2000 & 2000 & & pa & Panjabi & 106351 & 2000 & 2000 \\
    eo & Esperanto & 337074 & 2000 & 2000 & & pl & Polish & 999991 & 2000 & 2000 \\
    es & Spanish & 999996 & 2000 & 2000 & & ps & Pashto & 77765 & 2000 & 2000 \\
    et & Estonian & 999993 & 2000 & 2000 & & pt & Portuguese & 999991 & 2000 & 2000 \\
    eu & Basque & 999991 & 2000 & 2000 & & ro & Romanian & 999986 & 2000 & 2000 \\
    fa & Persian & 999990 & 2000 & 2000 & & ru & Russian & 999982 & 2000 & 2000 \\
    fi & Finnish & 999993 & 2000 & 2000 & & rw & Kinyarwanda & 173028 & 2000 & 2000 \\
    fr & French & 999997 & 2000 & 2000 & & se & Northern Sami & 35207 & 2000 & 2000 \\
    fy & Western Frisian & 53381 & 2000 & 2000 & & sh & Serbo-Croatian & 267159 & 2000 & 2000 \\
    ga & Irish & 289339 & 2000 & 2000 & & si & Sinhala & 979052 & 2000 & 2000 \\
    gd & Gaelic & 15689 & 1605 & 1606 & & sk & Slovak & 999978 & 2000 & 2000 \\
    gl & Galician & 515318 & 2000 & 2000 & & sl & Slovenian & 999987 & 2000 & 2000 \\
    gu & Gujarati & 317723 & 2000 & 2000 & & sq & Albanian & 999971 & 2000 & 2000 \\
    ha & Hausa & 97980 & 2000 & 2000 & & sr & Serbian & 999988 & 2000 & 2000 \\
    he & Hebrew & 999973 & 2000 & 2000 & & sv & Swedish & 999972 & 2000 & 2000 \\
    hi & Hindi & 534192 & 2000 & 2000 & & ta & Tamil & 226495 & 2000 & 2000 \\
    hr & Croatian & 999991 & 2000 & 2000 & & te & Telugu & 63657 & 2000 & 2000 \\
    hu & Hungarian & 999982 & 2000 & 2000 & & tg & Tajik & 193744 & 2000 & 2000 \\
    id & Indonesian & 999976 & 2000 & 2000 & & th & Thai & 999972 & 2000 & 2000 \\
    ig & Igbo & 16640 & 1843 & 1843 & & tk & Turkmen & 11305 & 1852 & 1852 \\
    is & Icelandic & 999980 & 2000 & 2000 & & tr & Turkish & 999959 & 2000 & 2000 \\
    it & Italian & 999988 & 2000 & 2000 & & tt & Tatar & 100779 & 2000 & 2000 \\
    ja & Japanese & 999986 & 2000 & 2000 & & ug & Uighur & 72160 & 2000 & 2000 \\
    ka & Georgian & 377259 & 2000 & 2000 & & uk & Ukrainian & 999963 & 2000 & 2000 \\
    kk & Kazakh & 79132 & 2000 & 2000 & & ur & Urdu & 753753 & 2000 & 2000 \\
    km & Central Khmer & 110924 & 2000 & 2000 & & uz & Uzbek & 172885 & 2000 & 2000 \\
    kn & Kannada & 14303 & 917 & 918 & & vi & Vietnamese & 999960 & 2000 & 2000 \\
    ko & Korean & 999989 & 2000 & 2000 & & wa & Walloon & 103329 & 2000 & 2000 \\
    ku & Kurdish & 143823 & 2000 & 2000 & & xh & Xhosa  & 439612 & 2000 & 2000 \\
    ky & Kyrgyz & 25991 & 2000 & 2000 & & yi & Yiddish & 13331 & 2000 & 2000 \\
    li & Limburgan & 23780 & 2000 & 2000 & & zh & Chinese & 1000000 & 2000 & 2000 \\
    lt & Lithuanian & 999973 & 2000 & 2000 & & zu & Zulu & 36947 & 2000 & 2000 \\
    \bottomrule
    \end{tabular}
    \caption{Statistics of the OPUS-100 dataset after de-duplicating against the test set.}
    \label{tab:opus100}
\end{table*}

\subsection{Statistics of the Oracle Data}
\label{appen:oracle}
We concatenated all available parallel dev set individually for each target language.
Tables~\ref{tab:wmt_oracle} and \ref{tab:opus_oracle} illustrate the statistics of our oracle data, before and after excluding the parallel English-free data of 6 evaluation pairs.

\begin{table*}[h]
    \centering
    \begin{tabular}{ccc}
    \toprule
    Language & \#Oracle & \#Oracle$^*$ \\
    \midrule
    En   & 15976 & 15976\\
    Fr   & 8776 & 7180\\
    Cs   & 8776 & 7978\\
    De   & 8776 & 6383\\
    Fi   & 8776 & 8776\\
    Lv   & 8776 & 8776\\
    Et   & 8776 & 7180\\
    Ro   & 8776 & 7180\\
    Hi   & 8776 & 8776\\
    Tr   & 8776 & 7978\\
    Gu   & 8776 & 7978\\
    \bottomrule
    \end{tabular}
    \caption{Statistics of the constructed oracle data on WMT-10. Oracle$^*$ denotes after excluding the direct data of all evaluated language pairs. Language ranked by the available bilingual resources.}
    \label{tab:wmt_oracle}
\end{table*}

\begin{table*}[h]
    \centering
    \begin{tabular}{ccc}
    \toprule
    Language & \#Oracle & \#Oracle$^*$ \\
    \midrule
    Ar & 9600 & 6400 \\
    De & 9600 & 6400 \\
    En   & 148427 & 148427 \\
    Fr & 9600 & 6400 \\
    Gd & 1284 & 1284\\
    Ig & 1474 & 1474\\
    Kn & 733 & 733\\
    Nl & 9600 & 6400\\
    Or & 1053 & 1053\\
    Ru & 9600 & 6400\\
    Tk & 1481 & 1481\\
    Zh & 9600 & 6400\\
    Others & 1600 & 1600 \\
    \bottomrule
    \end{tabular}
    \caption{Statistics of the constructed oracle data on OPUS-100. Oracle$^*$ denotes after excluding the direct data of all evaluated language pairs. Language ranked alphabetically. ``Others'' includes all other 83 languages.}
    \label{tab:opus_oracle}
\end{table*}

\end{document}